\documentclass[10pt,twocolumn,letterpaper]{article}

\usepackage{multirow}
\usepackage{pifont}
\usepackage{booktabs}
\usepackage{todonotes}
\usepackage{amsmath}
\usepackage{bm}
\usepackage{graphics}
\usepackage{makecell}
\usepackage[accsupp]{axessibility} 

\usepackage[pagenumbers]{cvpr} 

\newcommand{\myparagraph}[1]{\vspace{0.2cm}\noindent\textbf{#1}.\ }
\newcommand{\architecture}{{BiSeNetFormer}}

\definecolor{cvprblue}{rgb}{0.21,0.49,0.74}

\usepackage[pagebackref,breaklinks,colorlinks,citecolor=cvprblue]{hyperref}

\def\paperID{27} 
\def\confName{CVPR}
\def\confYear{2024}

\title{The revenge of BiSeNet: Efficient Multi-Task Image Segmentation}

\author{Gabriele Rosi$^{1,2}$, Claudia Cuttano$^1$, Niccolò Cavagnero$^1$, Giuseppe Averta$^{1,2}$, Fabio Cermelli$^2$ \\
$^1$ Politecnico di Torino, $^2$ Focoos AI\\
$^1$ {\tt\small name.surname@polito.it}, $^2$ {\tt\small name.surname@focoos.ai}
}

\begin{document}
\maketitle
\begin{abstract}
Recent advancements in image segmentation have focused on enhancing the efficiency of the models to meet the demands of real-time applications, especially on edge devices. However, existing research has primarily concentrated on single-task settings, especially on semantic segmentation, leading to redundant efforts and specialized architectures for different tasks.
To address this limitation, we propose a novel architecture for efficient multi-task image segmentation, capable of handling various segmentation tasks without sacrificing efficiency or accuracy. We introduce \architecture, that leverages the efficiency of two-stream semantic segmentation architectures and it extends them into a mask classification framework. Our approach maintains the efficient spatial and context paths to capture detailed and semantic information, respectively, while leveraging an efficient transformed-based segmentation head that computes the binary masks and class probabilities.
By seamlessly supporting multiple tasks, namely semantic and panoptic segmentation, \architecture\ offers a versatile solution for multi-task segmentation. We evaluate our approach on popular datasets, Cityscapes and ADE20K, demonstrating impressive inference speeds while maintaining competitive accuracy compared to state-of-the-art architectures. Our results indicate that \architecture\ represents a significant advancement towards fast, efficient, and multi-task segmentation networks, bridging the gap between model efficiency and task adaptability.

\end{abstract}    
\section{Introduction}
\label{sec:intro}

\begin{figure}[th!]
    \centering
    \includegraphics[width=0.85\linewidth]{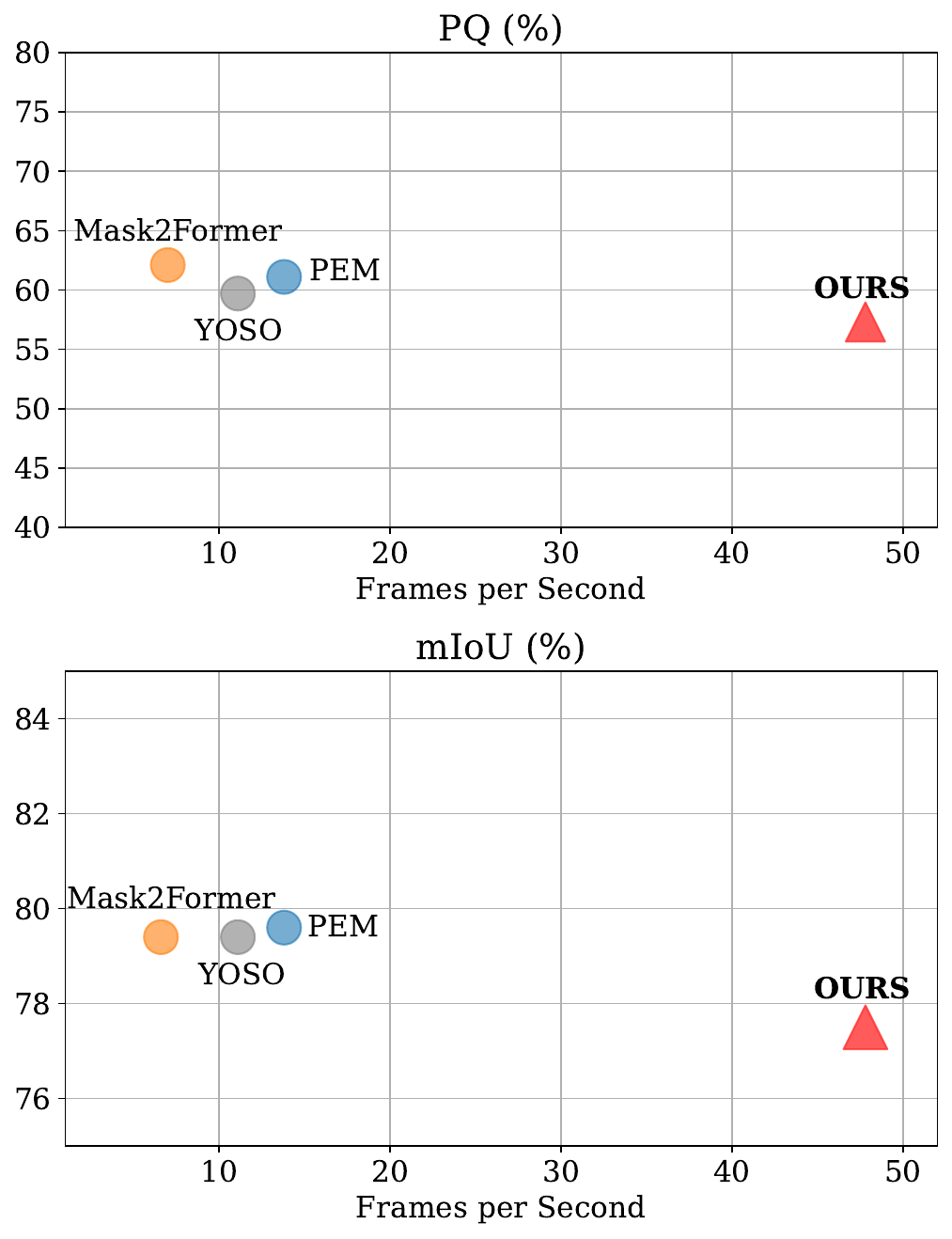}
    \caption{\architecture\ delivers comparable or superior performance in comparison to existing methods while being the fastest multi-task architecture for image segmentation.} \vspace{-11pt}
    \label{fig:teaser}
\end{figure}

In computer vision, image segmentation is a fundamental task, whose goal is to assign a class to each pixel in the image. 
Image segmentation can be categorized into several distinct settings. Semantic segmentation focuses on assigning class labels to each pixel (\eg road or person). Instance segmentation extends semantic segmentation with the capability to discriminate between different instances of the same countable class (``things'') but it disregards uncountable elements (``stuff'') like road or grass. Finally, panoptic segmentation attempts to retain the benefits of both settings providing instance-wise segmentation of ``things" while also segmenting ``stuff" classes.

The versatility of segmentation models makes them essential for a variety of downstream tasks, ranging from self-driving cars, to autonomous robots, augmented reality and surveillance. 
Since these applications require fast inference speed and low latency, especially on edge devices, the last decade has seen a growing interest in enhancing the efficiency of image segmentation networks while preserving the model accuracy \cite{pidnet, pp, stdc, bisenet, ddrnet}. However, despite the similarity among different image segmentation tasks, traditional approaches have evolved following different directions for each setting, providing optimized architectures that can only be operated on the specific scenario they were designed for, triplicating the research efforts.

For this reason, recently, increasing interest has been devoted towards the improvement of the capabilities of architectures beyond single-task settings, enabling multi-task image segmentation \cite{maskformer, m2f, kmax}. In particular, MaskFormer \cite{maskformer} proposed a new segmentation paradigm, named mask classification, able to seamlessly address all the different image segmentation tasks.
Nevertheless, existing mask classification models rely on computationally intensive modules, hindering real-time performance. Developing multi-task efficient models for segmentation is a highly under-explored topic, even though this direction offers a promising avenue for creating networks that excel in both inference speed and multi-task adaptability, without requiring modifications.

To achieve this goal, we discuss two distinct strategies. The first is to optimize the efficiency of mask classification networks. Adopting tailored architectural refinements, real-time performance can be attained even moving directly from mask classification approaches, as demonstrated by the success of YOSO \cite{yoso} and PEM \cite{cavagnero2024pem}, which significantly accelerated MaskFormer \cite{maskformer} while still improving its results in terms of accuracy.
The second strategy would instead tackle the problem from an opposite direction, seeking a solution to intrinsically redesign efficient two-stream semantic segmentation models in a mask-based framework. We argue that, since these architectures are intrinsically more efficiency-aware than the original MaskFormer \cite{maskformer}, working following the first intuition would promote model accuracy over efficiency, while the second strategy boosts efficiency more than performance, as clearly shown in \cref{fig:teaser}. 

With the aim of delivering the fastest multi-task architecture for image segmentation, in this paper we follow the second approach and we study a reliable technical solution to rethink two-stream architectures within a mask-classification framework. 
With this goal, building upon popular two-stream semantic segmentation architectures \cite{bisenet, bisenetv2, stdc}, we propose \architecture. It maintains the efficient two-stream design: a spatial path extracts high-resolution low-level details from the image, while a context path generates highly semantic visual features. To perform mask classification, we adopt a transformer decoder component that efficiently computes a set of segment embeddings leveraging the low-resolution context path features. These embeddings are then employed to compute a set of pairs composed by a binary mask and the respective class probabilities, which compose the segmentation output.
Thanks to its design, \architecture\ seamlessly supports multiple tasks, such as semantic and panoptic segmentation.

We evaluate \architecture\ on both tasks on two popular datasets, Cityscapes \cite{cityscapes} and ADE20K \cite{ade}, comparing inference speed and accuracy of our architecture with state-of-the-art multi-task and task-specific models. \architecture\ achieves impressive speed on all the benchmarks while showcasing comparable performance to both task-specific networks and slower multi-task architectures. In addition, we conducted tests using a low-end GPU (NVIDIA T4) and an edge device (NVIDIA Jetson Orin). Notably, our method maintains consistent inference speed even on resource-constrained hardware, affirming its suitability for real-world deployment.

Our paper contributes with: 
\begin{itemize}
    \item we propose \architecture, that redesigns efficient two-branch semantic segmentation architecture to operate in multiple segmentation tasks, 
    \item we demonstrate through an extensive experimental validation that \architecture\ showcases outstanding inference speed (\ie up to 100 FPS) while obtaining performance close to existing slower multi-task architectures.
\end{itemize}

\begin{figure*}[t]
    \centering
    \includegraphics[width=0.88\linewidth]{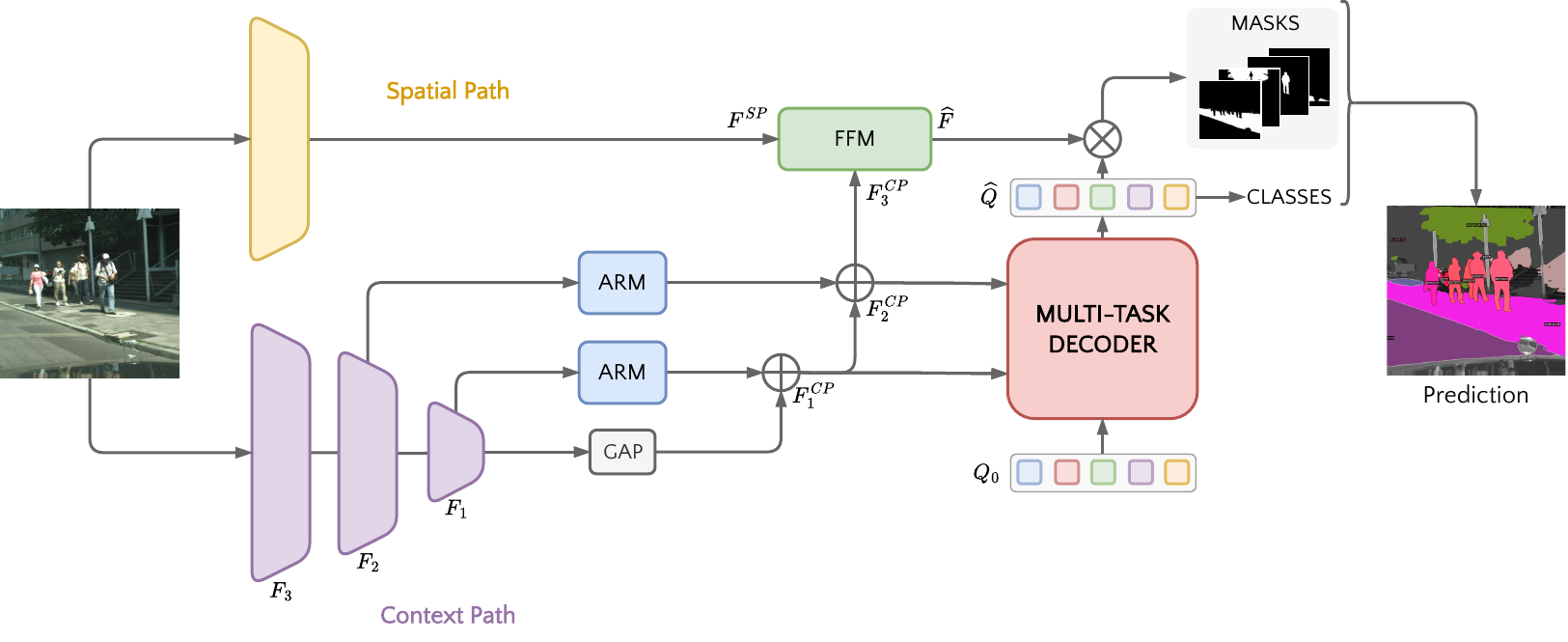}
    \caption{\textbf{Architecture of \architecture} with the three main components highlighted: spatial path (yellow), context path (violet) and transformer decoder (red). The spatial path extracts high-resolution features from the input image; the context path enlarges the receptive field and obtains highly semantical visual features; the transformer decoder takes as input a set of learnable queries and the high-resolution features to produce segment embeddings. A segmentation head then merges the spatial and context path features and then computes the final binary masks and class probabilities.} \vspace{-10pt}
    \label{fig:method}
\end{figure*}

\section{Related Works}
\label{related}

\myparagraph{Image Segmentation}
Recent advancements in computer vision aim to develop architectures capable of handling diverse image segmentation tasks without requiring modifications to loss functions or components. DETR \cite{detr} pioneered this approach, demonstrating successful object detection and panoptic segmentation using an end-to-end set prediction framework based on mask classification. Building upon this foundation, MaskFormer \cite{maskformer} proposed a specialized architecture based on the mask classification approach, establishing new benchmarks for both semantic and panoptic segmentation.
Mask2Former \cite{m2f} further enhanced results and convergence speed, achieving superior performance against both general-purpose and task-specific architectures. More recently, kMaX-DeepLab \cite{kmax} explored an alternative to traditional attention mechanisms through k-Means clustering operation. 
While these innovations have driven significant improvements in multi-task segmentation, the resource-intensive nature of these models presents a significant hurdle for real-world deployment on edge devices with limited computational capabilities.

\myparagraph{Efficient Image Segmentation}
Significant effort has been devoted to reduce the computational demands of image segmentation models, leading to efficient architectures suitable for real-time deployment \cite{bisenet, stdc, pidnet, wang2020solo, pandeeplab, realtimepan, yoso}. However, these endeavors predominantly addressed specialized architectures tailored for a single segmentation task.
In semantic segmentation, BiSeNet \cite{bisenet} proposed a two-branch model to separately process contextual information and spatial details, which are then fused together by a dedicated module to yield the prediction. STDC \cite{stdc} further built upon BiSeNet proposing a more efficient structure by reducing redundant architectural components. DDRNet \cite{ddrnet} introduced bilateral connections among the two branches, leading to a more effective information fusion. PIDNet \cite{pidnet} extended this framework with a three-branch design to improve object boundary detection. Panoptic segmentation approaches like UPSNet \cite{upsnet} merged instance and segmentation predictions to yield the panoptic output, while FPSNet \cite{fpsnet} focused on attention-based mask merging. Other techniques employed object localization via points or boxes \cite{lpsnet,pandeeplab,realtimepan}. YOSO \cite{yoso} recently demonstrated an efficient transformer-based method for both ``thing'' and ``stuff'' mask predictions.
While these specialized solutions excel in their domains, their task-specific nature fragments research efforts. PEM \cite{cavagnero2024pem} addressed this by proposing a real-time architecture for both semantic and panoptic segmentation. Although PEM \cite{cavagnero2024pem} and YOSO \cite{yoso} offer a valuable trade-off between performance and speed, there is still a significant gap between their inference speed and the one of specialized architectures.
In contrast, this work presents a highly efficient network that prioritizes speed while maintaining the flexibility to operate across multiple image segmentation tasks. Our approach aims to surpass existing specialized architectures in efficiency, addressing the limitations of both task-specific and recent multi-task solutions.

\section{\architecture}
\label{sec:method}

Semantic segmentation two-stream architectures \cite{bisenet, bisenetv2, stdc, ddrnet} excel in performance and speed. However, their design presents obstacles in extending them beyond semantic segmentation. The core limitation lies in the per-pixel classification head, which can inherently perform semantic segmentation only due to its static output structure (\ie a fixed set of output classes) and its inability in predicting variable instance counts.
To address this issue, we propose \architecture, a new solution to adapt two-branch architectures for mask classification, enabling them to handle multiple segmentation tasks, \eg semantic and panoptic segmentation. In the following, we first recall how mask classification works for segmentation and we then describe our solution in detail.

\subsection{Mask-classification Framework}
Recently, MaskFormer \cite{maskformer} has proposed a paradigm shift in image segmentation, unifying all the different segmentation tasks under a single approach, named mask classification. The idea is to perform segmentation in two steps: (1) divide the image into N different segments (where $N$ may be different from the number of classes), and (2) associate a class label to each segment. One of the major strengths of mask-based approaches is their versatility, allowing them to seamlessly address different segmentation tasks, since the same class can be associated with a variable number of masks.
Formally, given an image, $\bm{I} \in \mathbb{R}^{3 \times H \times W}$, the mask-classification head outputs a set of $N$ binary masks $\bm{M} \in \{0, 1\}^{N \times H \times W}$ each associated with a probability distribution $\bm{p}_i \in \Delta^{K+1}$, where $(H, W)$ is the height and width of the image, and $K+1$ is the number of classes, plus an additional ``\textit{no object}'' class.

\subsection{\architecture\ architecture}
While the first implementations of mask classification models \cite{maskformer, m2f, kmax} delivered an impressive performance, their architectural design often relied on compute-intensive modules. This design choice strongly limits their suitability for real-time applications and their deployment on resource-limited edge devices, where processing speed is critical. More recent works \cite{yoso, cavagnero2024pem} sought to address this efficiency bottleneck, achieving both performance improvements and increased efficiency. However, the design choices used to improve models efficiency in \cite{yoso, cavagnero2024pem} still promote accuracy rather than computational demand, which experimentally results in limited inference speed.

In contrast, we propose a two-stream architecture that prioritizes inference speed while integrating mask classification capabilities, pushing the limits of efficiency to previously unmatched capabilities for multi-task architectures. 
Our solution is composed of a \textit{Spatial Path} and a \textit{Context Path} inspired by BiSeNet \cite{bisenet}, followed by a transformer-based segmentation head \cite{m2f} to compute the output binary masks and class probabilities. The overall architecture is illustrated in \cref{fig:method}.

\myparagraph{Spatial path} The \textit{Spatial path} is responsible for the preservation of the spatial information of the image, enabling precise segmentation masks. Input images are processed sequentially, progressively reducing the spatial resolution up to $\frac{1}{8}$ of the initial image size. As confirmed by previous works \cite{bisenet, chen2017deeplab, chen2017rethinking}, $\frac{1}{8}$ offers an excellent trade-off between model accuracy and speed.
Formally, the spatial path takes as input an image $\bm{I}$ and generates a high-resolution feature $\bm{F}^{SP} \in \mathbb{R}^{C \times \frac{H}{8} \times \frac{W}{8}}$. 

\myparagraph{Context path} Conversely, the \textit{Context path} focuses on enriching each pixel representation by providing broader contextual information.
To this end, this module aims to enlarge the receptive field while avoiding an excessive increase in computational demand. 
In particular, it first extracts two low-resolution features $\bm{F}_1$ and $\bm{F}_2$, respectively being $\frac{1}{32}$ and $\frac{1}{16}$ of the original image. Such features encode high-level semantic information which is crucial to properly contextualize each pixel. A global average pooling is then applied on $\bm{F}_1$ to extract a feature that encodes a global scene understanding. Then, the features are recombined and upsampled to obtain higher resolution representations with relevant semantic information. 

Formally, the context path outputs three features $\bm{F}^{CP}_1$, $\bm{F}^{CP}_2$, $\bm{F}^{CP}_3$ with resolution equal to $\frac{1}{32}$, $\frac{1}{16}$, and $\frac{1}{8}$ of the original image, respectively. They are computed as follows: 
\begin{align}
    \centering
    \label{eq:sum_4}
    \bm{F}^{CP}_1 &= \texttt{ARM}(\bm{F}_1) + \texttt{GAP}(\bm{F}_1),\\
    \bm{F}^{CP}_2 &= \texttt{ARM}(\bm{F}_2) + \texttt{Up}(\bm{F}^{CP}_1),\\
    \bm{F}^{CP}_3 &= \texttt{Up}(\bm{F}^{CP}_2),
    \label{eq:sum_3}
\end{align}
where \texttt{GAP} denotes a global average pooling layer, \texttt{Up} an upsampling operation, and \texttt{ARM} refers to the Attention Refinement Module, proposed in \cite{bisenet}.

\myparagraph{Transformer decoder}
The goal of the transformer decoder is to generate $N$ segment embeddings $\widehat{\bm{Q}} \in \mathbb{R}^{N \times D}$, which are used to compute the final binary masks and probability distributions. To obtain favorable segmentation results, the segment embeddings should properly encode a representation of the classes (or instances) present in the image. For this reason, they are computed refining a set of $N$ learnable queries $\bm{Q}_0 \in \mathbb{R}^{N \times D}$ with the multi-scale features $\bm{F}^{CP}_1$ and $\bm{F}^{CP}_2$ coming from the Context Path.

To iteratively refine the queries, we employ a stack of Mask2Former \cite{m2f} transformer blocks. This block strategically integrates masked cross-attentions (MCAs), self-attentions (SAs), and feed-forward networks (FFNs). Masked cross-attention selectively aligns each query with relevant image features extracted from the Context Path. Self-attention allows queries to interact and learn contextual relationships between each other, improving the learned representations. Lastly, FFN is a 2-layer feed-forward network adopted to introduce additional non-linearity for complex pattern recognition.
Formally, given the query at the previous stage $\bm{Q}_{i-1}$ and a context path feature $\bm{F}^{CP}_j$, the transformer block computes the following operations:
\begin{align}
    \centering
    \bm{Q}_{i}' &= \texttt{MCA}(\bm{Q}_{i-1}, \bm{F}^{CP}_j, \bm{\mathcal{M}}_{i-1}) + \bm{Q}_{i-1},\\
    \bm{Q}_{i}'' &= \texttt{SA}(\bm{Q}_{i}') + \bm{Q}_{i}',\\
    \bm{Q}_{i} &= \texttt{FFN}(\bm{Q}_{i}'') + \bm{Q}_{i}'',
    \label{eq:transformer_decoder}
\end{align}
where $\bm{\mathcal{M}}_{i-1}$ is a binary mask obtained from the binarized output of the previous decoder layer, similarly to \cite{m2f}.

We sequentially apply this transformer block on the multi-scale features $\bm{F}^{CP}_1$ and $\bm{F}^{CP}_2$ for $L$ times. We consider $\widehat{\bm{Q}}$ as the output query of the last transformer block. Note that, to prioritize computational efficiency, we limit masked cross-attention to $\bm{F}^{CP}_1$ and $\bm{F}^{CP}_2$, avoiding the use of the highest resolution features $\bm{F}^{CP}_3$, which would introduce a consistently larger computational overhead. 

\myparagraph{Segmentation Head}
The goal of the segmentation head is to generate the final predictions, \ie a set of binary masks each associated with a class probability.

The class probabilities $\{\bm{p}_i \in \Delta^{K+1}\}_{i=1}^{N}$ are directly obtained by applying a classification layer, parameterized with $\bm{W}_{CLS} \in \mathbb{R}^{D \times K+1}$, over each segment embeddings $\widehat{\bm{Q}}$, followed by a softmax activation. Formally, denoting the $i$ embedding as $\widehat{\bm{Q}}_i$, we compute:
\begin{equation}
    \bm{p}_i = \texttt{softmax}(\widehat{\bm{Q}}_i \cdot \bm{W}_{CLS}).
\end{equation}

To obtain the final binary segmentation masks, instead, a two-steps procedure is required. First, high-resolution features coming from the Spatial ($\bm{F}^{SP}$) and Context ($\bm{F}_3^{CP}$) Paths are fused together; then, we multiply the resulting feature map and the segment embeddings $\widehat{\bm{Q}}$ to generate the final prediction. The fusion of the features coming from the spatial and context paths is a crucial point, because the former focuses on low-level visual details while the latter extracts high-level semantic concepts. The fusion operation is performed following \cite{bisenet} and employing a Feature Fusion Module (FFM). It first concatenates $\bm{F}^{SP}$ and $\bm{F}_3^{CP}$ to obtain a feature which retains both spatial and context information. Then, to obtain the final output feature $\widehat{\bm{F}} \in \mathbb{R}^{C \times \frac{H}{8} \times \frac{W}{8}}$, a reweighting strategy similar to SENet \cite{se} is employed to facilitate feature selection and combination. Formally, we compute
\begin{align}
      \bm{F}_{cat} &= \texttt{CONCAT}(\bm{F}^{SP},\bm{F}_3^{CP}), \\
      \bm{F}_{cat}' &= \texttt{CONV}_{3\times3}(\bm{F}_{cat}), \\
      \bm{F}_{avg} &= \texttt{sigmoid}(\texttt{FFN}(\texttt{GAP}(\bm{F}_{cat}'))),\\
      \widehat{\bm{F}} &= \bm{F}_{cat}' * \bm{F}_{avg} + \bm{F}_{cat}',
\end{align}
where \texttt{CONCAT} is a concatenation operation, $\texttt{CONV}_{3\times3}$ is a $3\times3$ convolution followed by ReLU and Batch Normalization, \texttt{GAP} is a global average pooling operator, \texttt{FFN} is a two-layers ReLU feed-forward network followed by a \texttt{sigmoid} function and $*$ is a broadcast element-wise multiplication.

Finally, the feature $\widehat{\bm{F}}$ is combined with the segment embeddings $\widehat{\bm{Q}}$ to obtain the final binary masks. First, a feed-forward network with two hidden layers converts the segment embeddings into $N$ mask embeddings $\widehat{\bm{M}} \in \mathbb{R}^{N \times C}$, having the same channel dimension $C$ of $\widehat{\bm{F}}$. Then, the binary masks are obtained via dot product between the features and the embeddings, followed by sigmoid activation. Formally, a pixel identified by $(h,w)$ of the output mask $m_i$ is computed as follows:
\begin{equation}
    m_i[h, w] = \texttt{sigmoid}(\widehat{\bm{M}}[i, :] \cdot \widehat{\bm{F}}[:, h, w]).
\end{equation}

\section{Experiments}
\label{sec:experiments}

To assess the validity of our method, we evaluate it on standard benchmarks for semantic and panoptic segmentation. We compare our method against both real-time and state-of-the-art architectures in terms of performances, latency and computational complexity.

\myparagraph{Datasets} We evaluate \architecture\ on two common datasets for semantic and panoptic segmentation: Cityscapes \cite{cityscapes} and ADE20K \cite{ade}.

The Cityscapes dataset features high-resolution images ($1024 \times 2048$ pixels) of urban street scenes taken from an egocentric perspective. The dataset is partitioned into a training set (2975 images), a validation set (500 images), and a testing set (1525 images). Image annotations span 19 distinct object and region classes.

The ADE20K dataset is composed by 20,000 training images and 2,000 validation images, featuring diverse locations and a wide variety of objects. The dataset includes images of varying sizes.

\myparagraph{Evaluation metrics} We evaluate the inference speed of our proposed architecture using frames per second (FPS) on a NVIDIA V100 GPU as performance metric. To measure semantic segmentation performance, we adopt the standard mean Intersection over Union (mIoU) \cite{mIoU}. For the more comprehensive panoptic segmentation, the Panoptic Quality (PQ) metric \cite{pq} is employed. PQ combines Segmentation Quality (how well segments align, measured by IoU) and Recognition Quality (how accurately objects are classified). To gain further insights, we also report PQ separately for ``thing'' classes (PQ$_{th}$) and ``stuff'' classes (PQ$_{st}$).

\subsection{Implementation details}
\myparagraph{Architecture} To reduce the computation on the Context Path we follow the implementation of \cite{stdc}, using the output of the spatial path as its input. In the following, we will refer to the union of the two paths as \textit{backbone}.
Unless explicitly stated, all the employed backbones are pretrained on ImageNet-1k \cite{imagenet}. Our architectural configuration features 3 transformer stages with two blocks each, a hidden dimension of 256 and 8 attention heads. Feed-forward networks adopt an expansion factor of four, while the number of object queries is set to 100. Furthermore, we project all the output features from the Context Path to a common dimensional space of size 128.

\myparagraph{Training settings} Our models are trained with AdamW \cite{adamw} optimizer and a step learning rate scheduler. The learning rate is set at 0.0003 and weight decay to 0.05 for both datasets. A learning rate multiplier of 0.1 is applied to the backbone. Our models are trained for 90k and 160k iterations on Cityscapes and ADE20K, respectively, with a batch size of 32. 
During training and inference, different crop sizes have been adopted depending on the dataset and the task. More specifically, we adopted a fixed crop size of $1024 \times 1024$ during training on the Cityscapes dataset for both tasks. At inference time, the whole image is employed. Differently, during training on ADE20K, we employed a fixed crop size of $512 \times 512$ for semantic segmentation and a fixed crop size of $640 \times 640$ for panoptic segmentation. During inference, the shorter side of the image is resized to fit the corresponding crop size.

\myparagraph{Losses} We follow Mask2Former \cite{m2f} with respect to the loss function adopted to train the networks. In particular, a binary cross-entropy loss is used to supervise the classification head. For accurate mask generation, instead, we use a combination of binary cross-entropy loss and dice loss \cite{dice}: $\mathcal{L}_{mask} = \lambda_{ce}\mathcal{L}_{ce} + \lambda_{dice}\mathcal{L}_{dice}$. The final loss is a combination of mask loss and classification loss: $\mathcal{L} = \mathcal{L}_{mask} + \lambda_{cls}\mathcal{L}_{cls}$. We set $\lambda_{ce} = 5.0$, $\lambda_{dice} = 5.0$ and $\lambda_{cls} = 2.0$ following the original implementation \cite{m2f}.

\begin{table}[t]
\vspace{-0.3cm}
\setlength{\tabcolsep}{1pt}
\centering
\begin{tabular}{@{}l|c|c|crr@{}}
\toprule
Method                       & Backbone        &  mIoU                & FPS                  & FLOPs             & Params     \\ \midrule
\multicolumn{6}{c}{Task-specific Architectures} \\ \midrule
BiSeNetV1 \cite{bisenet}     & R18           & 74.8                 & 65.5$^\dagger$       & 55G               & 49.0M    \\
BiSeNetV2-L\cite{bisenetv2}    & -               & 75.8                 & 47.3$^\ddagger$      & 119G              & -        \\
STDC1-Seg75 \cite{stdc}      & STDC1           & 74.5                 & 74.8$^\dagger$       & -                    & -       \\
STDC2-Seg75 \cite{stdc}      & STDC2           & 77.0                 & 58.2$^\dagger$       & -                    & -       \\
SegFormer-B0 \cite{xie2021segformer}         & -               & 76.2                 & 15.2                 & 125G                    & 3.8M                             \\
DDRNet-23-S \cite{ddrnet}    & -               & 77.8     & 108.1              & 36G      & 5.7M     \\
DDRNet-23 \cite{ddrnet}      & -               & 79.5     & 51.4               & 143G     & 20.1M    \\
PIDNet-S \cite{pidnet}       & -               & 78.8     & 93.2               & 46G      & 7.6M     \\
PIDNet-M \cite{pidnet}       & -               & 80.1     & 39.8               & 197G     & 34.4M    \\
PIDNet-L \cite{pidnet}       & -               & 80.9     & 31.1               & 276G     & 36.9M    \\ \midrule
\multicolumn{6}{c}{Multi-task Architectures} \\ \midrule
MaskFormer \cite{maskformer}       & R101      & 78.5     & 10.1               & 559G     & 60.2M       \\
Mask2Former \cite{m2f}       & R50             & 79.4     & 6.6                & 523G     & 44.0M        \\
YOSO \cite{yoso}             & R50             & 79.4     & 11.1               & 268G     & 42.6M        \\ 
PEM       & STDC1           & 78.3   & 24.3    & 92G     & 17.0M    \\
PEM & STDC2           & 79.0   & 22.0    & 118G    & 21.0M    \\
\midrule
\textbf{\architecture}       & STDC1   & 75.4   & 59.0    & 56G & 13.1M    \\
\textbf{\architecture}       & STDC2   & 77.5   &  47.8   & 82G & 17.1M    \\
\bottomrule
\end{tabular}
\caption{\textbf{Semantic segmentation on Cityscapes with 19 categories.} $\dagger$: resolution of 1536x768. $\ddagger$: resolution of 1024x512.}\vspace{-5pt}
\label{tab:cit_sem}
\end{table}
\subsection{Results}

\myparagraph{Semantic segmentation on Cityscapes} We compare \architecture\ with state-of-the-art methods for semantic segmentation on Cityscapes \cite{cityscapes} dataset in \cref{tab:cit_sem}. \architecture\ achieves 77.5 mIoU at 47.8 FPS. Compared to other multi-task architectures, \architecture\ achieves impressive results with nearly four times the inference speed and only a limited performance drop. Compared to task-specific models, \architecture\ outperforms other two-branches architectures like BiSeNet \cite{bisenet,bisenetv2} and STDC \cite{stdc} in terms of overall accuracy, while a direct comparison in terms of inference speed cannot be made since the latters use a smaller image size for testing (see \cref{tab:cit_sem}). DDRNet \cite{ddrnet} and PIDNet \cite{pidnet}, instead, deliver slightly higher mIoU and FPS results, which however come at the cost of an inherent limitation: DDRNet has been tailored for this task and their end-to-end design strongly limits modular changes, like the use of different backbones or segmentation heads.

\myparagraph{Semantic segmentation on ADE20K}
We present a comprehensive performance evaluation of \architecture\ against established state-of-the-art methods for semantic segmentation on ADE20K dataset \cite{ade} in \cref{tab:ade_sem}. \architecture\ shows exceptional capabilities in balancing accuracy with computational efficiency, achieving a noteworthy inference speed of 99.4 FPS. Compared to both task-specific and multi-task architectures, \architecture\ demonstrates significant performance gains. Notably, our solution even outperforms task-specific approaches, like BiSeNet \cite{bisenet} by a substantial margin of 10 mIoU and the bigger PIDNet \cite{pidnet} by 4.4 mIoU. These findings emphasize the proficiency of \architecture\ in addressing the complex challenge of accurately distinguishing among the 150 distinct classes found within the ADE20K dataset. 

\begin{table}[t]
\vspace{-0.3cm}
\setlength{\tabcolsep}{1.7pt}
\centering
\begin{tabular}{@{}l|c|c|crr@{}}
\toprule
Method        & Backbone &  mIoU  & FPS     & FLOPs    & Params \\ \midrule
\multicolumn{6}{c}{Task-specific Architectures}                 \\ \midrule
BiSeNetV1 \cite{bisenet}      & R18    &  35.1      &  143.1      &  15G   & 13.3M       \\
BiSeNetV2-L \cite{bisenetv2}  & -        &  28.5      &  106.7      &  12G   & 3.5M       \\
STDC1 \cite{stdc}             & STDC1    &  37.4      &  116.1      &  8G    & 8.3M       \\
STDC2 \cite{stdc}             & STDC2    &  39.6      &  78.5       &  11G   & 12.3M       \\
SegFormer-B0 \cite{xie2021segformer}          & -        &  37.4      &  50.5       &  8G   & 4.8M       \\
DDRNet-23-S \cite{ddrnet}     & -        &  36.3      & 96.2        & 4G     & 5.8M       \\
DDRNet-23 \cite{ddrnet}       & -        &  39.6      & 94.6        & 18G    & 20.3M      \\
PIDNet-S \cite{pidnet}        & -        &  34.8      & 73.5        & 6G     & 7.8M       \\
PIDNet-M \cite{pidnet}        & -        &  38.8      & 73.3        & 22G    & 28.8M       \\
PIDNet-L \cite{pidnet}        & -        &  40.5      & 65.4        & 34G    & 37.4M       \\ \midrule
\multicolumn{6}{c}{Multi-task Architectures}                     \\ \midrule
MaskFormer \cite{maskformer}  & R50      & 44.5   & 29.7        &  55G        & 41.3M       \\ 
Mask2Former \cite{m2f}        & R50      & 47.2   & 21.5        & 70G         &  44.0M      \\
YOSO \cite{yoso}              & R50      & 44.7   & 35.3       &  37G        &  42.0M      \\ 
PEM       & STDC1  & 39.6   &  43.6   & 16G   & 17.0M    \\
PEM        & STDC2  & 45.0   &  36.3       &  19G   &  21.0M   \\ 
\midrule
\textbf{\architecture}        & STDC1    & 42.1   & 112.9   & 8G    & 13.1M\\
\textbf{\architecture}        & STDC2    & 44.9   & 99.7    &  11G   &  17.2M

\\\bottomrule

\end{tabular}
\caption{\textbf{Semantic segmentation on ADE20K with 150 categories.} FLOPs are measured at resolution 512x512.}\vspace{-10pt}
\label{tab:ade_sem}
\end{table}

\myparagraph{Panoptic segmentation on Cityscapes} The results for panoptic segmentation on Cityscapes \cite{cityscapes} dataset are reported in \cref{tab:cit_pan}.
Compared with all competitor methods, \architecture\ achieves a notable speed of 22.3 FPS when equipped with ResNet50 - the fastest among all competitors - while obtaining a panoptic quality of 57.3. Paying a price of a very limited performance drop, \architecture\ proved to be surprisingly fast, enabling an inference speed two times faster than the fastest competitor. When equipped with the STDC2 backbone, the inference speed is nearly four times higher, while achieving a Panoptic Quality score of 57.3.

\myparagraph{Panoptic segmentation on ADE20K} \cref{tab:ade_pan} presents the results for panoptic segmentation on the ADE20K \cite{ade} dataset.
In this last case, \architecture\ shows a non-negligible performance drop of approximately 8 points w.r.t. competitors \cite{yoso, cavagnero2024pem}, but it showcases an impressive inference speed of 99.7 FPS and 77.4 FPS when equipped with STDC2 and ResNet50, respectively. Moreover, our architecture shows the lowest number of parameters and the lowest computational complexity among all the other approaches. This performance-speed trade-off warrants further investigation to determine the underlying causes and potentially inform optimizations to \architecture.

\begin{table}[t]
\resizebox{\columnwidth}{!}
{
\setlength{\tabcolsep}{2pt}
\centering
\begin{tabular}{@{}l|c|ccc|crr@{}}
\toprule
Method                            & Backbone   & PQ        & PQ$_{th}$ & PQ$_{st}$ & FPS            & FLOPs   & Params    \\ \midrule
Mask2Former \cite{m2f}            & R50        & 62.1      & -         & -         & 4.1            & 519G    & 44.0M     \\
UPSNet \cite{upsnet}              & R50        & 59.3      & 54.6      & 62.7      & 7.5            &  -      &  -        \\
LPSNet \cite{lpsnet}              & R50        & 59.7      & 54.0      & 63.9      & 7.7            &   -     &  -        \\
PanDeepLab \cite{pandeeplab}      & R50        & 59.7      & -         & -         & 8.5            &  -      &  -        \\
FPSNet \cite{fpsnet}              & R50        & 55.1      & -         & -         & 8.8$^\dagger$  &  -      &  -        \\
RealTimePan \cite{realtimepan}    & R50        & 58.8      & 52.1      & 63.7      & 10.1           &  -      & -         \\
YOSO \cite{yoso}                  & R50        & 59.7      & 51.0      & 66.1      & 11.1           & 265G    & 42.6M     \\
PEM                               & R50        & 61.1      & 54.3      & 66.1      & 13.8           & 237G    & 35.6M \\
\midrule
\textbf{\architecture} & STDC2    & 57.3  & 48.6 & 66.0 & 47.8 & 83G & 17.1M    \\ 
\textbf{\architecture} & R50      & 57.5 & 52.2 & 62.4  & 22.3 & 199G & 31.7M    \\ \bottomrule
\end{tabular}}
\caption{\textbf{Panoptic segmentation on Cityscapes with 19 categories.} $\dagger$: measured on a Titan GPU.}
\label{tab:cit_pan}
\end{table}
\begin{table}[t]
\centering
\resizebox{\columnwidth}{!}
{
\setlength{\tabcolsep}{2pt}
{\begin{tabular}{@{}l|c|ccc|ccc@{}}
\toprule
Method                        & Backbone   & PQ   & PQ$_{th}$ & PQ$_{st}$ & FPS   & FLOPs & Params \\ \midrule
BGRNet \cite{bgrnet}          & R50   & 31.8 & 34.1      & 27.3      & -     & -    & - \\
MaskFormer \cite{maskformer}  & R50   & 34.7 & 32.2      & 39.7      & 29.7  & 86G  & 45.0M    \\
Mask2Former \cite{m2f}        & R50   & 39.7 & 39.0      & 40.9      & 19.5  & 103G & 44.0M \\
kMaxDeepLab \cite{kmax}       & R50   & 42.3 & -         & -         & -     & -    & - \\
YOSO \cite{yoso}              & R50   & 38.0 & 37.3      & 39.4      & 35.4  & 52G  & 42.0M \\ 
PEM                           & R50   & 38.5  & 37.0 & 41.1      &  35.7 & 47G & 35.6M   \\ \midrule
\textbf{\architecture}        & STDC2 & 30.8  & 29.3  & 34.1  & 99.7  & 17G & 17.2M   \\
\textbf{\architecture}        & R50   & 31.6  & 30.2  & 34.6  & 77.4  & 39G & 31.6M   \\ 
\bottomrule
\end{tabular}}}
\caption{\textbf{Panoptic segmentation on ADE20k with 150 categories.} FLOPs are measured at resolution 640x640.} 
\label{tab:ade_pan}
\end{table}

\begin{table}[t]
\centering
\begin{tabular}{c|ccc}
\toprule
\multirow{2}{*}{Backbone}        & \multicolumn{3}{c}{\textit{FPS}} \\
              & V100 & T4 & Jetson ORIN  \\ \midrule
\multicolumn{4}{c}{\textit{ADE20K Panoptic [640 $\times$ 640]}} \\ \midrule
STDC1           & 112.9 & 79.5   &  38.2    \\
STDC2           & 99.7 & 65.5   & 35.2    \\
R50             & 77.4 & 33.8   & 33.8   \\ \midrule
\multicolumn{4}{c}{\textit{Cityscapes Panoptic [1024 $\times$ 2048]}} \\ \midrule
STDC1    & 59.0 & 24.0  &  19.8     \\
STDC2    & 47.8 & 18.9   &  16.5   \\
R50      & 37.6  & 7.7  & 9.7    \\
\bottomrule
\end{tabular}
\caption{FPS of \architecture\ on different devices in panoptic segmentation.} \vspace{-10pt}
\label{tab:deploy}
\end{table}

\subsection{Deployment on different hardware}
\cref{tab:deploy} showcases \architecture's remarkable adaptability across a spectrum of diverse Nvidia devices: V100, T4 and Jetson ORIN. Considering the ADE20K \cite{ade} dataset in the panoptic segmentation settings, the architecture achieves real-time speed on powerful V100 GPUs with all the backbone in analysis. Of particular note is its strong performance on the T4, reaching 79.5 FPS, which well positions it for applications where dedicated GPUs are available but with less computational power and energy consumption than a V100.
Perhaps most impressively, \architecture\ demonstrates its true potential for edge deployment on the Jetson ORIN. Achieving a frame rate of 38.2 FPS on the ADE20K dataset, it unlocks the possibility of real-time panoptic segmentation in resource-constrained environments. Similar trends are observed with the Cityscapes dataset. While the overall frame rates are expectedly lower due to the increased image resolution, \architecture\ continues to exhibit impressive speeds across different devices and backbones.

\begin{table}[t]
\centering
\begin{tabular}{c|c|ccc}
\toprule
Resolutions                  & \begin{tabular}[c]{@{}c@{}}\# Stages\end{tabular} & PQ & FPS & Latency \\ \midrule
\multirow{3}{*}{$\bm{F}^{CP}_1$, $\bm{F}^{CP}_2$}     & 1  & 54.8  &  61.0   & 16.4        \\
                             & 2  &  55.1   &  53.0   & 18.9        \\
                             & \textbf{3}  &  \textbf{57.3}  & 47.8    &  20.9       \\ \midrule
\multirow{3}{*}{$\bm{F}^{CP}_1$, $\bm{F}^{CP}_2$, $\bm{F}^{CP}_3$} & 1  & 56.9   &  44.3   &  22.6       \\
                             & 2  & 57.4  &  33.8   &  33.8       \\
                             & 3  & \textbf{58.9}   &  27.1   &  36.9       \\ \bottomrule
\end{tabular}
\caption{Ablation on input resolutions and number of decoding stages on Cityscapes.}
\label{tab:abl_resolutions}
\end{table}

\begin{table}[t]
    \centering
    \begin{tabular}{c|ccc|c} 
    \toprule
     N & PQ   & PQ$_{th}$ & PQ$_{st}$ & FLOPs \\ 
    \midrule
    50  & 54.8  & 42.9  & 63.5  & 80G  \\
    100 & 57.3  & 48.6  & 66.0  & 83G  \\
    200 & 58.0  & 49.8  & 69.1  &  88G \\
    \bottomrule
    \end{tabular}
    \caption{Ablation on number of queries on Cityscapes.}
    \label{tab:abl_numqueries} \vspace{-13pt}
\end{table}

\begin{figure*}[t]
    \centering
\includegraphics[width=\textwidth]{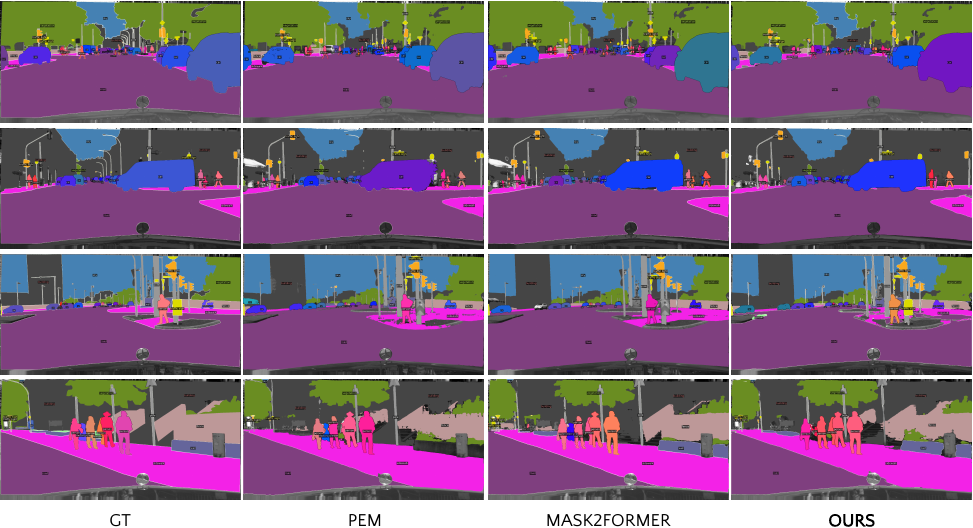} \vspace{-15pt}
    \caption{Qualitative Results for panoptic segmentation on the Cityscapes \cite{cityscapes} dataset. Best seen with colors and digital zoom.} 
    \label{fig:panoptic} \vspace{-10pt}
\end{figure*}

\subsection{Ablation study}

We perform ablation studies on the proposed architecture, in order to assess the contribution of each module to the final performance. All the ablations are performed on Cityscapes \cite{cityscapes} dataset in the panoptic setting.

\myparagraph{Input resolution and number of transformer decoder stages} 
\cref{tab:abl_resolutions} provides valuable insights into the relationship between input resolutions, transformer decoder stage configurations, and \architecture's overall performance. A key observation is that limiting the transformer decoder stages to two input features ($\bm{F}^{CP}_1$ and $\bm{F}^{CP}_2$) strikes an optimal equilibrium between accuracy and computational efficiency. This configuration not only matches the results obtained using all three features with two decoding layers but also delivers a substantial 14 FPS advantage in inference speed. 
These findings suggest that a careful selection of input features can have a profound impact on the efficiency of \architecture's transformer decoder stages without sacrificing accuracy. Based on this analysis, we opted for three decoder stages in the final \architecture\ configuration. This decision was motivated by its favorable performance across all evaluated tasks, even though higher FPS can be achieved with a smaller number of decoding stages.

\myparagraph{Number of queries} \cref{tab:abl_numqueries} reveals how varying the number of queries within transformer decoder influences \architecture's performance. Increasing the number of queries leads to improvements in PQ, particularly in the PQ$_{st}$ category. However, this comes at the cost of increased computational overhead as measured in FLOPs. Notably, the gains begin to taper off as the query count grows, suggesting a point of diminishing returns. Our decision to employ 100 queries represents a strategic balance between accuracy and efficiency, maximizing performance improvements while keeping computational complexity in check.

\subsection{Qualitative results}

\cref{fig:panoptic} demonstrates the qualitative performance of \architecture\ on the Cityscapes \cite{cityscapes} dataset for panoptic segmentation tasks. Visual comparisons against both resource-intensive models like Mask2Former \cite{m2f} and the lighter architectures PEM \cite{cavagnero2024pem} highlight that, despite being largely more efficient, \architecture\ achieve qualitatively similar results. It correctly recognizes all the class instances in the images, both the larger ones, such as the cars and person, as well as the smaller traffic signs and lights. Furthermore, it precisely segments their boundaries, obtaining segmentation masks similar to less efficient alternatives.

\section{Conclusions}
\label{sec:conclusions}
In this paper, we present \architecture, a novel architecture for multi-task image segmentation that combines the efficiency of two-stream semantic segmentation architectures with a mask-based classification approach. Our approach addresses the need for real-time, efficient, and adaptable segmentation networks capable of handling various tasks such as semantic and panoptic segmentation.
Through extensive experiments on Cityscapes and ADE20K datasets, we have demonstrated the effectiveness of \architecture\ in achieving impressive inference speeds while maintaining competitive accuracy compared to state-of-the-art architectures.
The success of \architecture\ highlights the importance of creating efficient architectures for multi-task scenarios, thereby reducing redundancy in research efforts and promoting the development of versatile and high-performance computer vision systems. Future work will focus on further refining \architecture\ and exploring additional tasks and datasets to expand its applicability and impact in real-world applications.

\footnotesize{
\myparagraph{Acknowledgements}
This study was carried out within the FAIR - Future Artificial Intelligence Research and received funding from the European Union Next-GenerationEU (PIANO NAZIONALE DI RIPRESA E RESILIENZA (PNRR) – MISSIONE 4 COMPONENTE 2, INVESTIMENTO 1.3 – D.D. 1555 11/10/2022, PE00000013). 
We also acknowledge the Sustainable Mobility Center (CNMS) which received funding from the European Union Next Generation EU (Piano Nazionale di Ripresa e Resilienza (PNRR), Missione 4 Componente 2 Investimento 1.4 "Potenziamento strutture di ricerca e creazione di "campioni nazionali di R\&S" su alcune Key Enabling Technologies") with grant agreement no. CN\_00000023. 
This manuscript reflects only the authors’ views and opinions, neither the European Union nor the European Commission can be considered responsible for them.}

{
    \small
    \bibliographystyle{ieeenat_fullname}
    \bibliography{main}
}

\end{document}